\documentclass{article}

\usepackage{microtype}
\usepackage{graphicx}
\usepackage{subfigure}
\usepackage{booktabs} 

\usepackage{cite}
\usepackage{graphicx}
\usepackage{latexsym}
\usepackage{amsmath}
\usepackage{amssymb}
\usepackage{subfigure}
\usepackage{xcolor}
\usepackage{enumerate}
\usepackage{comment}
\usepackage{subfigure}
\usepackage{multirow}
\usepackage{bm}
\usepackage{wrapfig}
\usepackage{times}

\usepackage{hyperref}

\usepackage[accepted]{icml2020}

\icmltitlerunning{DRWR: A Differentiable Renderer without Rendering for Unsupervised 3D Structure Learning from Silhouette Images}

\begin{document}

\twocolumn[
\icmltitle{DRWR: A Differentiable Renderer without Rendering for Unsupervised 3D Structure Learning from Silhouette Images}




\begin{icmlauthorlist}
\icmlauthor{Zhizhong Han}{to,goo}
\icmlauthor{Chao Chen}{to}
\icmlauthor{Yu-Shen Liu}{to}
\icmlauthor{Matthias Zwicker}{goo}
\end{icmlauthorlist}

\icmlaffiliation{to}{School of Software, BNRist, Tsinghua University, Beijing 100084, P. R. China}
\icmlaffiliation{goo}{Department of Computer Science, University of Maryland, College Park, USA}

\icmlcorrespondingauthor{Yu-Shen Liu}{liuyushen@tsinghua.edu.cn}

\icmlkeywords{Unsupervised Structure Learning, 3D Point Cloud, Differentiable Renderer, Single Image 3D Reconstruction, 3D Generative Model, Deep Learning Model}

\vskip 0.3in
]



\printAffiliationsAndNotice{}  

\begin{abstract}
Differentiable renderers have been used successfully for unsupervised 3D structure learning from 2D images because they can bridge the gap between 3D and 2D.
To optimize 3D shape parameters, current renderers rely on pixel-wise losses between rendered images of 3D reconstructions and ground truth images from corresponding viewpoints. Hence they require interpolation of the recovered 3D structure at each pixel, 
visibility handling, 
and optionally evaluating a shading model. In contrast, here we propose a \textit{Differentiable Renderer Without Rendering} (DRWR) that omits these steps. 
DRWR only relies on a simple but effective loss that evaluates how well the projections of reconstructed 3D point clouds cover the ground truth object silhouette. Specifically, DRWR employs a smooth silhouette loss
to pull the projection of each individual 3D point inside the object silhouette, and 
a structure-aware repulsion loss 
to push each pair of projections that fall inside the silhouette far away from each other. 
Although we omit surface interpolation, visibility handling, and shading, our results demonstrate that DRWR achieves state-of-the-art accuracies under widely used benchmarks, outperforming previous methods both qualitatively and quantitatively. In addition, our training times are significantly lower due to the simplicity of DRWR.
\end{abstract}

\section{Introduction}
Learning to represent and reconstruct 3D structure is a core problem in 3D computer vision. Supervised deep learning methods~\cite{cvprpoint2017,nipspoint17,MeschederNetworks,xu2019disn} have been highly successful by directly learning from 3D data provided as meshes, point clouds, or voxel volumes. However, these methods require large amounts of 3D data in training, which is expensive and time consuming to obtain. In contrast, unsupervised 3D structure learning, that is, 3D structure learning without 3D supervision, is an attractive and promising alternative because it requires only images as training data.

\begin{figure}[tb]
  \centering
   \includegraphics[width=\linewidth]{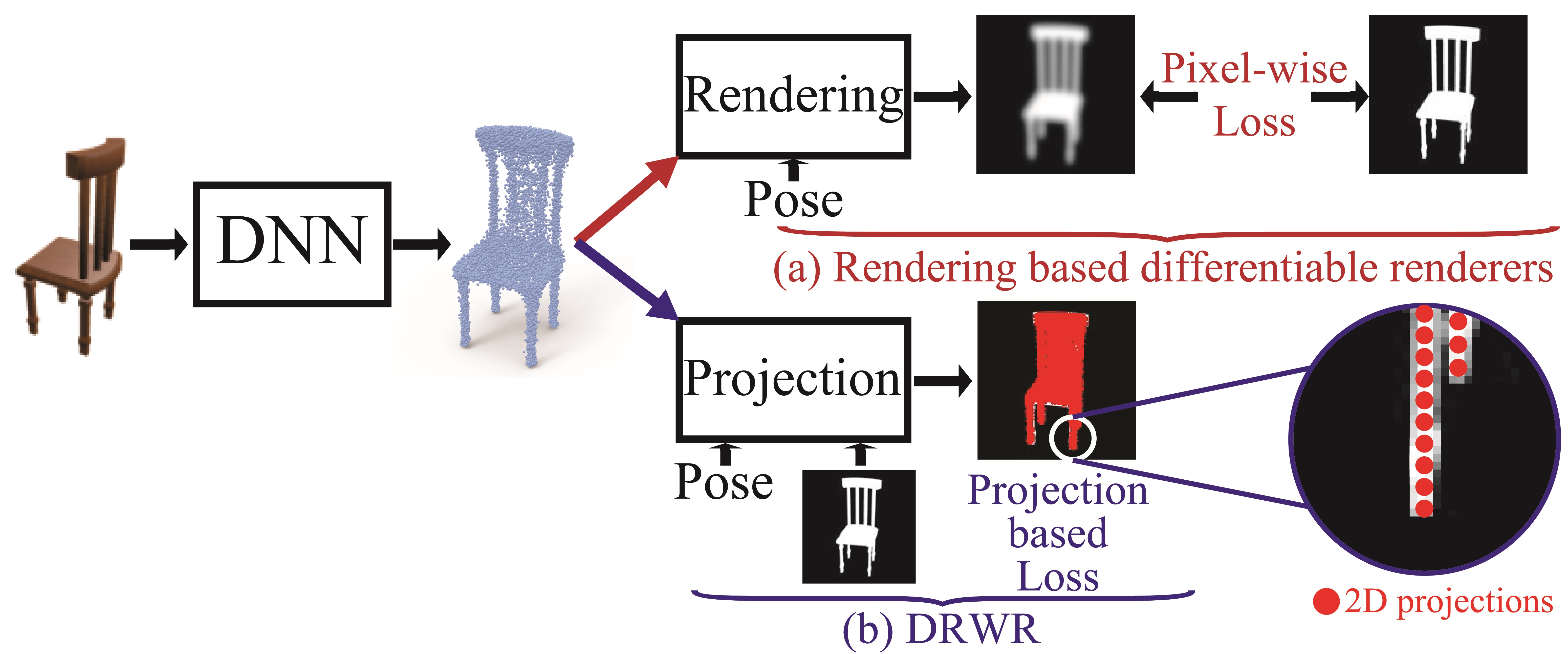}
  %
  %
\caption{\label{fig:difference} Illustration of the difference between rendering-based differentiable renderers with a pixel-wise loss in (a) and DRWR with a projection-based loss without rendering in (b). }
\end{figure}

Differentiable renderers~\cite{InsafutdinovD18,Navaneet2019,navaneet2019differ,Yifan:DSS:2019} are a core component of unsupervised 3D structure learning methods. They can bridge the gap between 3D to 2D by enabling the computation of gradients of 2D loss functions with respect to 3D structure. 2D loss functions are usually defined based on differences in RGB pixel values or pixel-wise silhouette coverage. By rendering a predicted 3D structure from a specific view angle into an image, and then evaluating a loss function based on the difference between the rendered and ground truth image, the parameters of deep learning models can be optimized to recover 3D structures that are consistent with the ground truth image, as illustrated in Fig.~\ref{fig:difference}(a). To evaluate pixel-wise loss functions, previous techniques render images using some form of interpolation of the recovered 3D structure at each pixel, such as rasterization, visibility handling (e.g., z-buffering), and optionally per-pixel shading. For point cloud reconstruction, differentiable renderers have been proposed based on rasterizing Gaussian functions into 3D grids~\cite{InsafutdinovD18} and on 2D planes~\cite{Navaneet2019,navaneet2019differ}, or using surface splatting~\cite{Yifan:DSS:2019}. While pixel-wise interpolation, visibility handling, and shading in these approaches significantly increase the computational cost, one of our main insights here is that these steps do not actually contribute to accurate 3D structure learning.


To show this, we propose a \textit{Differentiable Renderer Without Rendering} (DRWR) for unsupervised 3D point cloud reconstruction from 2D silhouette images. In contrast to pixel-wise losses in previous differentiable renderers, DRWR produces a loss only from the 2D projections of the 3D points, without pixel-wise interpolation, visibility handling, or shading, as shown in Fig.~\ref{fig:difference}(b). Intuitively, the DRWR loss captures how well the projected points cover the object silhouette, or the foreground. More specifically, the loss function includes a unary and a pairwise loss. The unary loss is designed to pull the projection of each 3D point into the foreground, while the pairwise loss pushes each pair of projections that lie inside the foreground far away from each other. This ensures that the entire foreground is covered and it prevents points from clumping. In addition, to avoid getting stuck in a local minima, we construct a smooth silhouette loss as the unary loss, which is designed to produce non-zero gradients for all points until their projections move into the foreground. To make the optimization more efficient and accurate, we also formulate the pairwise loss in a structure-aware manner, where we adaptively take into account the repulsion between each pair of projections only when both of them appear in the foreground. Our experiments show that DRWR outperforms all previous methods in achieving state-of-the-art results under widely used benchmarks. In summary, our main contributions are as follows:

\begin{enumerate}[i)]
\item We propose DRWR to justify the idea of conducting unsupervised 3D structure learning for point clouds using a differentiable renderer without rendering, that is, without pixel-wise interpolation of 3D structure, visibility handling, or shading.
\item We demonstrate that DRWR reduces training time while improving the state-of-the-art accuracy of reconstructed point clouds under widely used benchmarks.
\item We introduce a smooth silhouette loss and a structure-aware repulsion loss based on the projections of 3D points. The resulting model can be trained efficiently and robustly.
\end{enumerate}

\section{Related work}
Deep learning models have made significant progress in different 3D applications~\cite{nipspoint17,Park_2019_CVPR,MeschederNetworks,Zhizhong2016b,Zhizhong2016,Han2017,HanTIP18,Zhizhong2018seq,Zhizhong2019seq,parts4features19,3DViewGraph19,3D2SeqViews19,HanCyber17a,Hu2019Render4CompletionSM,hutaoaaai2020,MAPVAE19,seqxy2seqzeccv2020,p2seq18,l2g2019,wenxin_2020_CVPR}. Here, we briefly review differentiable renderers for different 3D representations including voxel grids, meshes, implicit functions and point clouds, which is most related to our methods.

\noindent\textbf{Voxel grids. }Yan et al.~\cite{YanNIPS2016} selected the maximum occupancy value along a ray to learn to reconstruct 3D voxel grids from silhouette images. Gadelha et al.~\cite{DBLP:conf/3dim/GadelhaMW17} employed orthogonal projection using simple projection function to bridge 3D voxel grids to silhouette images. Tulsiani et al.~\cite{TulsianiZEM17} derived a differentiable formulation by leveraging ray collision probabilities. These methods work with known camera poses. Then, Tulsiani et al.~\cite{mvcTulsiani18} extended ~\cite{TulsianiZEM17} with an additional network to simultaneously predict camera poses. Similarly, Gadelha et al. extended the projection~\cite{DBLP:conf/3dim/GadelhaMW17} in the presence of viewpoint uncertainties.

\noindent\textbf{Meshes. }OpenDR~\cite{Loper:ECCV:2014} is the pioneer of differentiable rendering based renderers, which approximates gradients with respect to pixel positions to back-propagate. With hand-crafted gradients, Kato et al.~\cite{KatoUH18} also achieved the adjustment of faces. To analytically compute gradients,~\cite{Liu:Paparazzi:2018} and~\cite{liu2018beyond} back-propagated the image
gradients to the face normals which are further used to update vertex positions via chain rule. Liu et al.~\cite{liu2019softras} introduced SoftRas with a probabilistic rasterization and assigned each pixel to all faces of a mesh. Similarly, Chen et al.~\cite{DBLP:journals/corr/abs-1908-01210} regarded rasterization as interpolation of local mesh properties by computing analytic gradients of foreground pixels.

\noindent\textbf{Implicit functions. }Implicit functions have been attracting more research interests as a new 3D shape representation to learn using deep learning models~\cite{pifuSHNMKL19,xu2019disn,MeschederNetworks,chen2018implicit_decoder,Oechsle_2019_CVPR,Park_2019_CVPR,DBLP:journals/corr/abs-1901-06802,Jiang2019SDFDiffDRcvpr}, due to its great representative ability with voxels or signed distance function in high resolutions. To reduce the computational cost on sampling implicit surface required in training, Vincent et al.~\cite{sitzmann2019srns} learned a mapping from world coordinates to a feature representation of local scene properties. Based on the idea of ray marching rendering, different differentiable renderers~\cite{DIST2019SDFRcvpr,Jiang2019SDFDiffDRcvpr,prior2019SDFRcvpr} were proposed to render the signed distance function. While Liu et al.~\cite{shichenNIPS} proposed a novel ray-based field probing technique to mine supervision for 3D occupancy fields. With the implicit differentiation,~\cite{Volumetric2019SDFRcvpr} derived analytically in a differentiable rendering formulation for implicit shape and texture representations.

\noindent\textbf{Point clouds. }Due to the compactness, point clouds are also an important 3D representation in deep learning based 3D shape understanding. However, this also brings an issue of sparseness among 2D projections of points. This issue makes the images with these projections impossible to directly compare with the ground truth images, which is hard to handle by differential renderers. To resolve this issue, different renderers mainly employed either dense points~\cite{lin2018learning} or different rendering approaches~\cite{InsafutdinovD18,Navaneet2019,navaneet2019differ,Yifan:DSS:2019} based on rasterization. Specifically, Lin et al.~\cite{lin2018learning} resorted to dense points and proposed pseudo-renderer to model the visibility using pooling. However, it is significantly affected by the number of points. Instead, rendering based methods~\cite{InsafutdinovD18,Navaneet2019,navaneet2019differ,Yifan:DSS:2019} approximated the distribution of point clouds using surface splatting~\cite{Yifan:DSS:2019} or gaussian functions in 3D space~\cite{InsafutdinovD18} and on 2D images~\cite{Navaneet2019,navaneet2019differ}. But rendering adds computational burden, while it is not clear whether it contributes to improving the reconstruction accuracy.

Different from these methods, DRWR can be used to generate point clouds containing arbitrary numbers of points and produces a loss for each point without rendering.

Moreover, DRWR is also much different from the previous methods~\cite{Dolphins6165306,Tulsiani7482798,abstractionTulsiani17} which did not leverage the rendering strategy to reveal the 3D structures from 2D images. These methods require strong priors, such as 3D template~\cite{Dolphins6165306,Tulsiani7482798} or primitives~\cite{abstractionTulsiani17}, and the guidance of 2D and 3D key point correspondence obtained by manually annotated~\cite{Dolphins6165306} or algorithm~\cite{Tulsiani7482798}. However, DRWR do not require any of these.


\begin{figure}[tb]
  \centering
   \includegraphics[width=\linewidth]{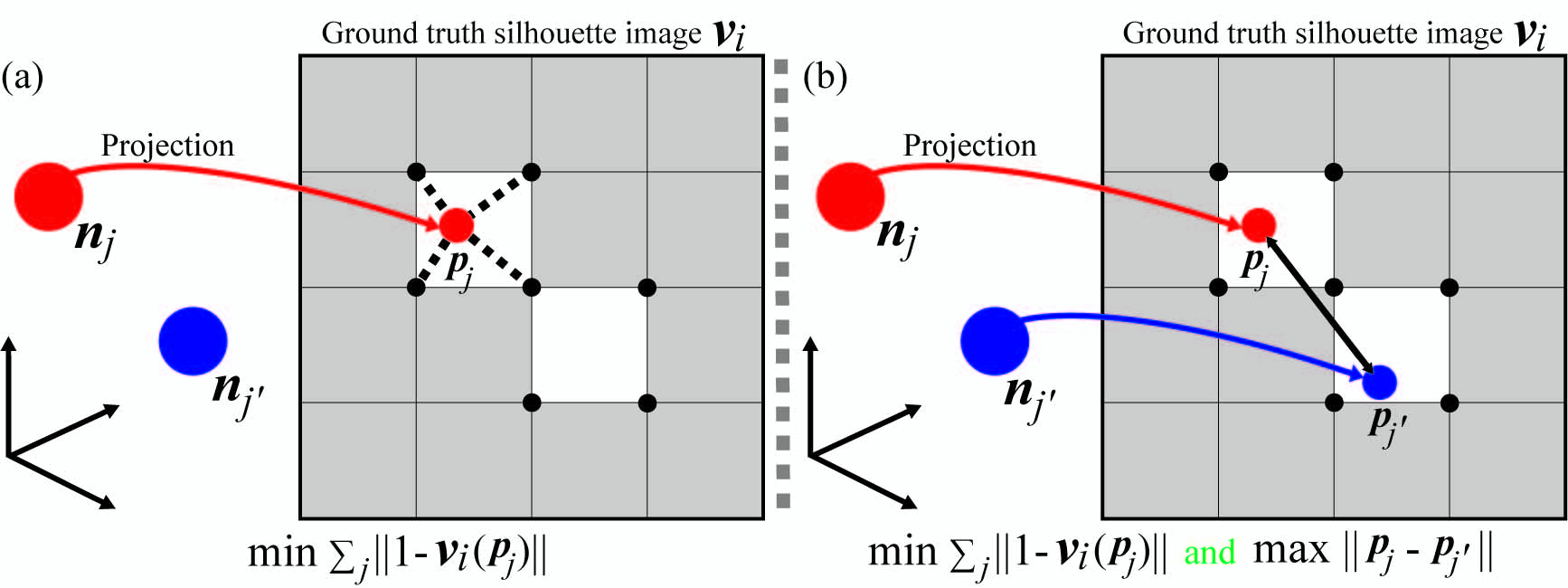}
  %
  %
\caption{\label{fig:frameworks} DRWR learns the structure of 3D point clouds from multiple silhouette images. For clarity, we show only two points $\bm{n}_j$ and $\bm{n}_{j^\prime}$ and one silhouette image $\bm{v}_i$. The goal of uniformly locating the projections $\bm{p}_j$ inside $\bm{v}_i$ is implemented by two losses: (a) the first loss pulls all projections into the foreground (white areas) by minimizing the error between 1 and the pixel value of each projection $\bm{v}_i(\bm{p}_j)$. The second loss in (b) pushes each pair of projections in the foreground far away from each other.}
\end{figure}

\section{Details of DRWR}
\noindent\textbf{Overview. }Our goal is to learn the structure of 3D point clouds $\bm{N}$ formed by $J$ points $\bm{n}_j$ only from $I$ ground truth silhouette images $\bm{v}_i$, where $j\in[1,J]$ and $i\in[1,I]$. Current differentiable renderers rely on rendering point clouds $\bm{N}$ into raster images $\bm{v}_i^\prime$ from the $i$-th view angle, which would then be used to produce a loss by comparing $\bm{v}_i^\prime$ and $\bm{v}_i$ pixel-by-pixel. We argue, however, that rendering with pixel-wise interpolation of 3D structure, visibility handling, and shading adds unnecessary computational cost, and accurate results can be achieved without these steps.

To demonstrate this, we propose DRWR, a differentiable renderer without rendering, providing a novel perspective for unsupervised learning of 3D structure. Denoting the projection of point $\bm{n}_j$ in view $i$ as $\bm{p}_j^i$, DRWR computes a loss by evaluating how well the sets of projected points $\{\bm{p}_j^i|j\in[1,J]\}$ cover the object silhouette. The DRWR loss consists of a unary and a pairwise term, as illustrated in Fig.~\ref{fig:frameworks}. Given a predicted 3D point cloud and a binary silhouette image, we compute the loss as follows (only two points for illustration): We first project the points $\bm{p}_j$ (short for $\bm{p}_j^i$) onto the silhouette images $\bm{v}_i$, denoting the pixel value of the projection $\bm{p}_j$ as $\bm{v}_i(\bm{p}_j)$. The unary loss penalizes points outside the foreground by simply computing the difference $1 - \bm{v}_i(\bm{p}_j)$, assuming that the foreground in the binary silhouette image has value $1$. Minimizing this loss will pull all projections into the foreground. In addition, the pairwise loss adjusts the spatial distribution of projected points by pushing pairs of projections in the foreground to be as far from each other as possible, as shown in Fig.~\ref{fig:frameworks}(b). DRWR adjusts the 3D locations of points $\bm{n}_j$ through their projections $\bm{p}_j$ by jointly optimizing these two losses. DRWR can produce this loss and its gradients for any generative neural network for 3D point clouds, enabling unsupervised training as shown in Fig.~\ref{fig:difference}(b).


\noindent\textbf{3D-to-2D Projection. }In our approach, we represent 3D point clouds $\bm{N}$ in an object centered coordinate system. Using the perspective transformation, we start by transforming the coordinate of each point $\bm{n}_j$ into the projection $\bm{p}_j^i$ on each silhouette image $\bm{v}_i$ from the $i$-th view angle. Using $\bm{T}_i$ as both extrinsic and intrinsic camera parameters of the $i$-th camera pose, we conduct the projection below,

\begin{equation}
\label{eq:transform}
[\bm{p}_j^i \quad 1]^T\sim\bm{T}_i[\bm{n}_j \quad 1]^T.
\end{equation}


\noindent\textbf{Smooth Silhouette Loss. }We model the unary loss as the difference between 1 and the pixel value $\bm{v}_i(\bm{p}_j^i)$ of each projection $\bm{p}_j^i$ on silhouette image $\bm{v}_i$. Here, we employ bilinear interpolation to obtain $\bm{v}_i(\bm{p}_j^i)$ using the binary pixel values of the nearest pixels around $\bm{p}_j^i$, as demonstrated in Fig.~\ref{fig:frameworks}(a). DRWR aims to pull all projections into the foreground on all silhouette images $\bm{v}_i$ by minimizing the loss
\begin{equation}
\label{eq:unary}
l_1(\bm{p}_j^i,\bm{v}_i)=\|1-\bm{v}_i(\bm{p}_j^i)\|.
\end{equation}
%

\begin{wrapfigure}{r}{0.5\linewidth}
\includegraphics[width=\linewidth]{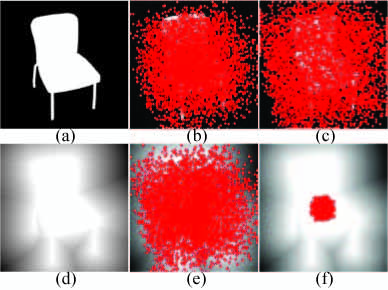}
\caption{\label{fig:localminimum} Visualization of the smooth silhouette loss. With binary pixel values in silhouette image $\bm{v}_i$ in (a), randomly initialized projections in (b) cannot be pulled into the foreground, as shown in (c). Using the smooth silhouette loss based on $\bm{v}_i^G$ in (d), the projections in (e) can be pulled into the foreground in (f).}
\end{wrapfigure}

However, we found that it is impossible to pull all the projections into the foreground by minimizing $l_1(\bm{p}_j^i,\bm{v}_i)$ in Eq.~(\ref{eq:unary}). As an example illustrated in Fig.~\ref{fig:localminimum}, we optimize a point cloud according to a silhouette image in Fig.~\ref{fig:localminimum}(a). Starting from randomly initialized points, whose projections are colored as red diamonds in Fig.~\ref{fig:localminimum}(b), the poorly optimized points are projected in Fig.~\ref{fig:localminimum}(c).



This problem occurs because the gradient of the loss in Eq.(\ref{eq:unary}) is merely from the pixel intensity difference between 1 and the pixel value $\bm{v}_i(\bm{p}_j^i)$ interpolated from the four nearby binary pixel values. This prohibits training if the projections $\bm{p}_j^i$ are far from the foreground, which would be a local minima. This issue also exists in motion estimation~\cite{Bergen92hierarchicalmodel-based,DBLP:conf/cvpr/ZhouBSL17,garg2016unsupervised,monodepth17}. Since these methods work on real images which contain large variability in texture, a widely used solution to resolve this issue is to employ an explicit multi-scale and smoothness loss that derives gradients from larger spatial regions directly. However, this solution cannot resolve our issue, since silhouette images have no texture variability in the background, which is still impossible to derive gradients even from larger spatial regions.


To resolve this issue, we introduce a smoothing procedure for the pixel values on the ground truth silhouette images $\bm{v}_i$ that we then use in the unary loss. The key idea behind our silhouette smoothing approach is to establish a progressively varying field in the background on $\bm{v}_i$, which leads to non-zero gradients anywhere in the background, while the pixel values in the foreground are not changed. We achieve this using the negative distance function as the smoothed values for each pixel $\bm{x}$ on the background in $\bm{v}_i$,
\begin{equation}
\bm{v}_i^G(\bm{x})=\left\{
\begin{array}{rcl}
1, & & {\bm{x}\in\Omega}\\
1-d(\bm{x},\partial\Omega),& & {\bm{x}\in\bar{\Omega}}\\
\end{array} \right.
\end{equation}
where $\Omega=\{\bm{x}|\bm{v}_i(\bm{x})=1\}$ is the foreground, $\bar{\Omega}=\{\bm{x}|\bm{v}_i(\bm{x})=0\}$ is the background, $\partial\Omega$ is the boundary of the foreground, and $d(\bm{x},\partial\Omega)$ is the L2 distance between $\bm{x}$ and its nearest $\partial\Omega$, which is normalized by the resolution of $\bm{v}_i$. We denote the smoothed silhouette images as $\bm{v}_i^G$ to distinguish them from the original silhouette images $\bm{v}_i$. Moreover, we normalize the smoothed pixel values in the background to lie in a range of 0 to 1 by minmax normalization, such that $\bm{v}_i^G(\bar{\Omega})=minmax(\bm{v}_i^G(\bar{\Omega}))$. Finally, based on Eq.~(\ref{eq:unary}), we define the smoothed silhouette loss as the following unary loss,
\begin{equation}
\label{eq:unary1}
l_1(\bm{p}_j^i,\bm{v}_i)=\|1-\bm{v}_i^G(\bm{p}_j^i)\|.
\end{equation}
We illustrate this loss using the example in Fig.~\ref{fig:localminimum}. With the smooth silhouette values in Fig.~\ref{fig:localminimum}(d), we can pull the projections of all randomly initialized points in Fig.~\ref{fig:localminimum}(e) into the foreground by minimizing Eq.~(\ref{eq:unary1}), as shown in Fig.~\ref{fig:localminimum}(f).



\noindent\textbf{Structure-aware Repulsion Loss. }The smooth silhouette loss is far from achieving our goal of uniformly locating projections in the foreground, as shown in Fig.~\ref{fig:localminimum}(f). To better cover the silhouette and more accurately capture the 3D shape, DRWR includes a pairwise loss to model the spatial relationship between each pair of projections. We design the pairwise loss such that minimizing it pushes projections inside the foreground far away from each other.

However, this pairwise loss is somewhat in conflict with the smooth silhouette loss, which tends to pull all projections close together, especially near the foreground boundaries.
In addition, near the foreground boundaries it is harder than deep inside the foreground to push two projections away from each other without pushing them into the background.
To resolve this issue, we propose a structure-aware repulsion loss as a pairwise term. This structure-aware repulsion loss adaptively weighs the repulsion between each pair of projections according to the structure around the projections. It increases repulsion for pairs of projections deep inside the foreground, reduces repulsion for pairs of projections around the foreground boundary, and cancels repulsion if any projection is in the background.


\begin{wrapfigure}{r}{0.3\linewidth}
\includegraphics[width=\linewidth]{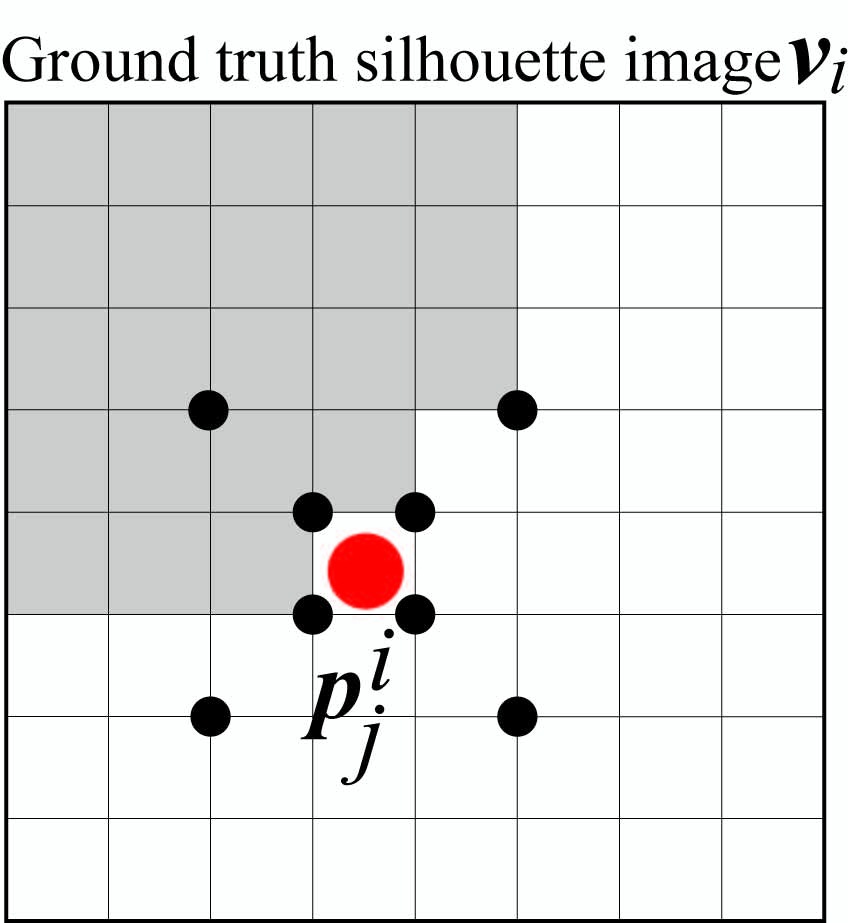}
\caption{\label{fig:weight} Multi-scale bilinear interpolation for boundary bias $\delta_{j}^i$.}
\end{wrapfigure}

Specifically, for each pair of projections $\bm{p}_j^i$ and $\bm{p}_{j^\prime}^i$, the L2 distance between them is $d(\bm{p}_j^i,\bm{p}_{j^\prime}^i)=\|\bm{p}_j^i-\bm{p}_{j^\prime}^i\|_2$, which we further normalize according to the resolution of the silhouette image. DRWR aims to maximize the distance $d(\bm{p}_j^i,\bm{p}_{j^\prime}^i)$, and we employ a Gaussian function to obtain a repulsion loss that decreases with increasing distance. Hence we can minimize the repulsion loss along with the smooth silhouette loss. For each projection $\bm{p}_j^i$, the structure-aware repulsion loss models its spatial relationships to all the other projections $\bm{p}_{j^\prime}^i$ as follows,
\begin{equation}
\label{eq:pairwise}
l_2(\bm{p}_j^i,\{\bm{p}_{j^\prime}^i\},\bm{v}_i)=\omega_j^i\sum_{j^\prime=1}^J \omega_{j^\prime}^i e^{(-d(\bm{p}_j^i,\bm{p}_{j^\prime}^i)/\sigma+\delta_{j}^i)},
\end{equation}
%


\begin{wrapfigure}{r}{0.5\linewidth}
\includegraphics[width=\linewidth]{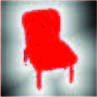}
\caption{\label{fig:final} Result with both losses.}
\end{wrapfigure}

where $\omega_{j}^i$ and $\omega_{j^\prime}^i$ are indicator weights for projections $\bm{p}_j^i$ and $\bm{p}_{j^\prime}^i$, $\sigma$ is the decay parameter, and $\delta_{j}^i$ is the boundary bias for projection $\bm{p}_{j}^i$. The indicator weight $\omega_{j}^i$ represents the degree to which projection $\bm{p}_j^i$ is located in the background. Small $\omega_{j}^i$ or $\omega_{j}^i=0$ decreases or completely removes the repulsion on $\bm{p}_j^i$, so that the smooth silhouette loss can pull $\bm{p}_j^i$ into the foreground immediately. The decay parameter $\sigma$ controls the range of repulsion. The boundary bias $\delta_{j}^i$ controls the distance to the foreground boundary where the repulsion on projection $\bm{p}_j^i$ is reduced.

We compute the indicator weight $\omega_{j}^i$ using bilinear interpolation from nearby binary pixel values on the silhouette image $\bm{v}_i$, such that $\omega_{j}^i=\bm{v}_i(\bm{p}_j^i)$, as shown in Fig.~\ref{fig:frameworks}(a). In addition, we employ a multi-scale bilinear interpolation approach to obtain the boundary bias $\delta_{j}^i$. We use binary pixel values from square neighborhoods at $R$ scales around projection $\bm{p}_j^i$ to conduct the interpolation, and take the mean of the interpolations from all $R$ scales as $\delta_{j}^i$. This is demonstrated in Fig.~\ref{fig:weight}, where $R=2$ scales around a projection (in red) are shown. In this example, some pixels (in black) on the two scales lie in the background (the shaded area), so the boundary bias $\delta_{j}^i$ is small, which accordingly reduces the repulsion on the projection. This approach enables DRWR to progressively decrease repulsion when $\bm{p}_j^i$ approaches the foreground boundary. As shown in Fig.~\ref{fig:final}, the structure-aware repulsion loss combined with the smooth silhouette loss successfully pulls all projections into the foreground (as in Fig.~\ref{fig:localminimum}(f)), while also uniformly covering the entire foreground area.

\noindent\textbf{Loss function. }DRWR minimizes the following overall loss function based on the two losses defined in Eq.~(\ref{eq:unary1}) and Eq.~(\ref{eq:pairwise}) to conduct unsupervised structure learning for 3D point clouds,
\begin{equation}
\label{eq:loss}
L=\frac{1}{I}\frac{1}{J}\sum_{i=1}^I\sum_{j=1}^J(l_1(\bm{p}_j^i,\bm{v}_i)+\beta l_2(\bm{p}_j^i,\{\bm{p}_{j^\prime}^i\},\bm{v}_i)),
\end{equation}
where $\beta$ is a weight to balance the two losses and the total loss $L$ averages over all $J$ points from all $I$ views.

\section{Experiments and Analysis}


\noindent\textbf{Datasets and metrics. } We conduct experiments involving 3D shapes in three categories from ShapeNet~\cite{ChangFGHHLSSSSX15}, including chairs, cars, and airplanes, which are commonly used for evaluation by our competitors. We also follow the same train/test splitting as in~\cite{TulsianiZEM17,InsafutdinovD18}, and we employ the five rendered views from each 3D shape and the ground truth point clouds released by~\cite{InsafutdinovD18}. Specifically, we employ rendered views at three different resolutions including $32^2$, $64^2$, and $128^2$, all corresponding to the same set of ground truth point clouds with different numbers of points.

We conduct numerical comparisons using the Chamfer distance (CD) between predicted and ground truth point clouds. For differentiable renderers for meshes or voxel grids, we also use volumetric IoU to conduct fair comparisons. Note that all reported CD or IoU values reported in our experiments are multiplied by 100 for better readability.

\noindent\textbf{Details. }For fair comparison, we employ exactly the same neural network architecture as the one introduced by Insafutdinov et al.~\cite{InsafutdinovD18}, however, replacing the  differentiable renderer~\cite{InsafutdinovD18} with DRWR. The approach by Insafutdinov et al.~\cite{InsafutdinovD18} implements structure learning of 3D point clouds using pairs of RGB images. For each pair, the network first generates a point cloud from the first RGB image, and then renders the predicted point cloud from the view angle of the second image. Their differentiable renderer produces a rendered silhouette image as its output, as shown in Fig.~\ref{fig:difference}(a), and the neural network is trained by minimizing the pixel-wise error between the rendered silhouette image and the silhouette of the second input image. In our approach, we replace their rendering-based differentiable renderer by DRWR, as shown in Fig.~\ref{fig:difference}(b). We omit rendering and simply leverage the projected positions of the generated point clouds to produce the loss and gradient required in the training. At test time, the trained network generates a 3D point cloud from a single RGB image.

We evaluate DRWR using this network with ground truth camera poses during projection. In addition, we employ RGB images with three different resolutions to train, and accordingly evaluate the generated point clouds in three resolutions including 2000, 8000, and 16000 points. We train the network using the Adam optimizer with a batch size of 16 rendered images (4 views of 4 shapes), where we iterate over $1\times10^{5}$ batches in each experiment.
%
%
%
%
We set $R=5$ to calculate the boundary bias $\delta_{j}^i$ for all projections during the optimization and set the decay parameter $\delta=1$ to decrease the repulsion between each pair of projections. We balance the two losses using $\beta=3$ in all our experiments.

\subsection{Comparison with the State-of-the-art}

\noindent\textbf{CD comparisons. }We first compare DRWR with rendering-based differentiable renderers in terms of CD. All compared renderers produce silhouettes of the predicted shapes to compute their loss with respect to ground truth silhouettes. We conduct the comparison by training the networks using silhouette images at three different resolutions as mentioned previously.

\begin{table*}[tb]
\centering
\caption{Comparison with point clouds renderers in terms of CD.}  
\resizebox{\linewidth}{!}{
    \begin{tabular}{|c|c|c|c|c|c|c|c|c|c|c|c|}  
     \hline
        & \multicolumn{5}{|c|}{Resolution 32 (2000)}  & \multicolumn{3}{|c|}{Resolution 64 (8000)} & \multicolumn{3}{|c|}{Resolution 128 (16000)} \\
       \hline
        & DRC & CAP & DPC-V & DPC &Ours& DPC-V & DPC& Ours & EPCG & DPC&Ours \\  
     \hline
     Plane &8.35&6.34&5.57&4.52&\textbf{4.01}&4.94&3.50&\textbf{3.18}&4.03&2.84&\textbf{2.66}\\
     Car&4.35&6.03&3.88&4.22&\textbf{3.81}&3.41&2.98&\textbf{2.89}&3.69&2.42&\textbf{2.40}\\
     Chair&8.01&6.11&5.57&5.10&\textbf{4.66}&4.80&4.15&\textbf{4.02}&5.62&3.62&\textbf{3.49}\\
     \hline
     Mean&6.90&6.16&5.01&4.61&\textbf{4.16}&4.39&3.55&\textbf{3.36}&4.45&2.96&\textbf{2.85}\\
     \hline
   \end{tabular}}
   \label{table:t10}
\end{table*}

We compare DRWR with Differentiable Ray Consistency (DRC)~\cite{TulsianiZEM17}, Efficient Point Cloud Generation (EPCG)~\cite{lin2018learning}, Continuous Approximation Projection (CAP)~\cite{Navaneet2019}, and Differentiable Point Clouds (DPC)~\cite{InsafutdinovD18}. The first renderer is voxel-based, and it is only available for voxel grids with a resolution of $32^3$ because of the cubic complexity of voxel grids. The other three renderers are point cloud-based, which is the same as DRWR.

We report the quantitative comparison in Table~\ref{table:t10}. Our results outperform all compared methods under all classes at all three resolutions. DRWR shows obvious advantages over voxel-based differentiable renderers including DRC and the voxel-based counterpart ``DPC-V'' of DPC, where DRWR recovers more geometry details in a more memory-efficient manner. In addition, by omitting rendering, DRWR also achieves
higher accuracies of the reconstructed point clouds compared to rendering-based differentiable point renderers, such as CAP, DPC, and EPCG. These results further demonstrate that DRWR is robust to changes in image resolutions and number of points.


\begin{wrapfigure}{r}{0.59\linewidth}
\includegraphics[width=\linewidth]{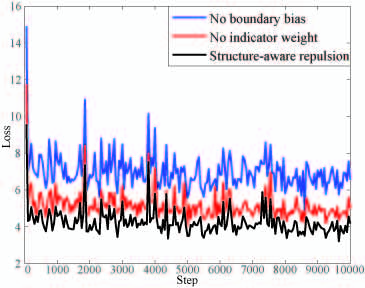}
\caption{\label{fig:final1} The efficiency of structure-aware repulsion.}
\end{wrapfigure}

\begin{figure*}[tb]
  \centering
   \includegraphics[width=\linewidth]{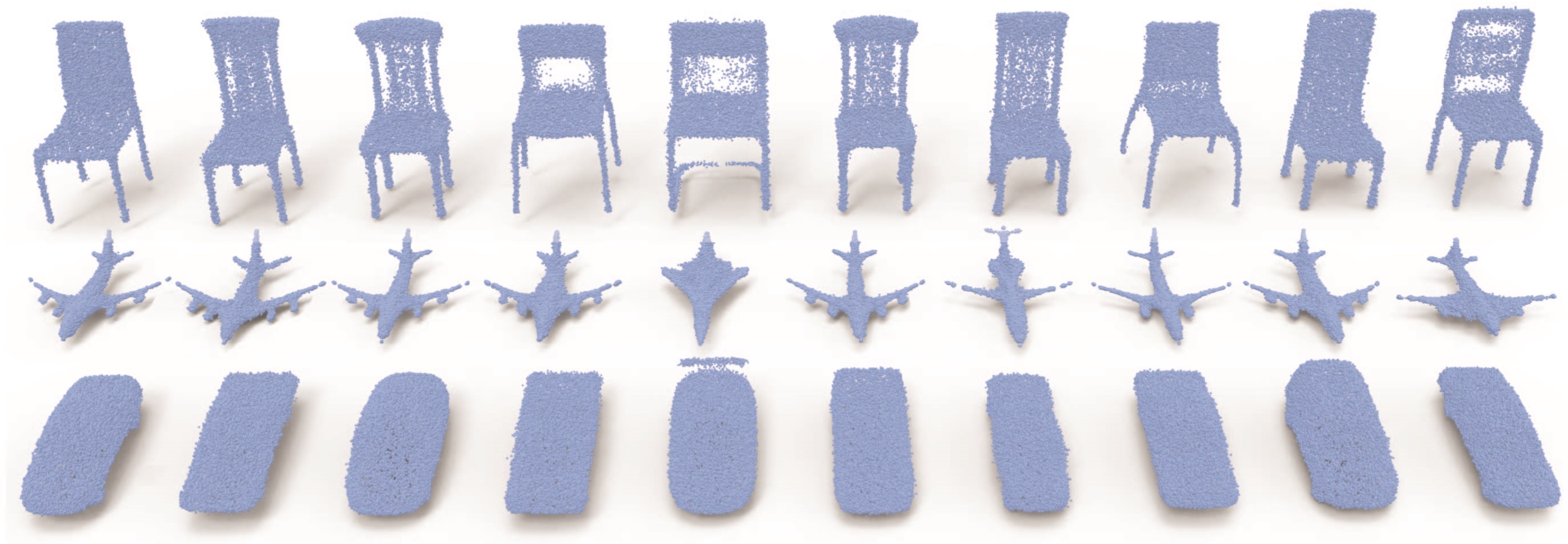}
  %
  %
\caption{\label{fig:Vis_128} Visualization of randomly selected shapes with $16000$ points in single image reconstruction in Table~\ref{table:t10}.}
\end{figure*}

\begin{figure}[tb]
  \centering
   \includegraphics[width=\linewidth]{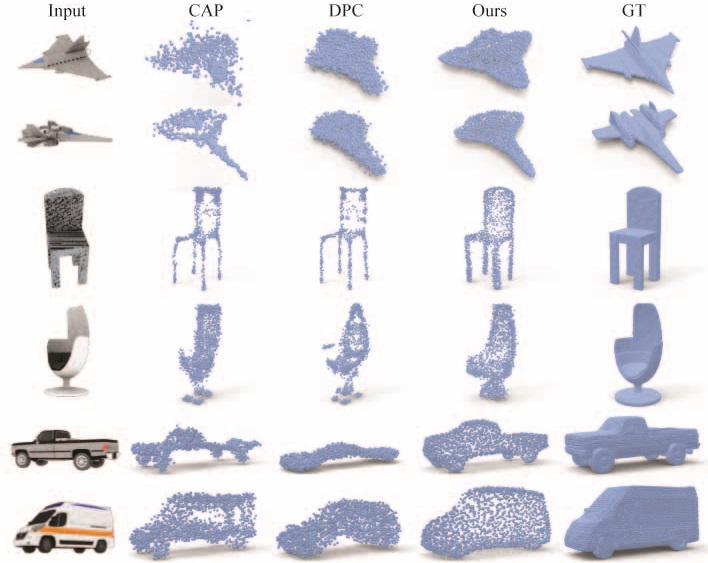}
  %
  %
\caption{\label{fig:PCDRComp} Qualitative comparison with differentiable renderers for point clouds.}
\end{figure}

\begin{figure}[tb]
  \centering
   \includegraphics[width=\linewidth]{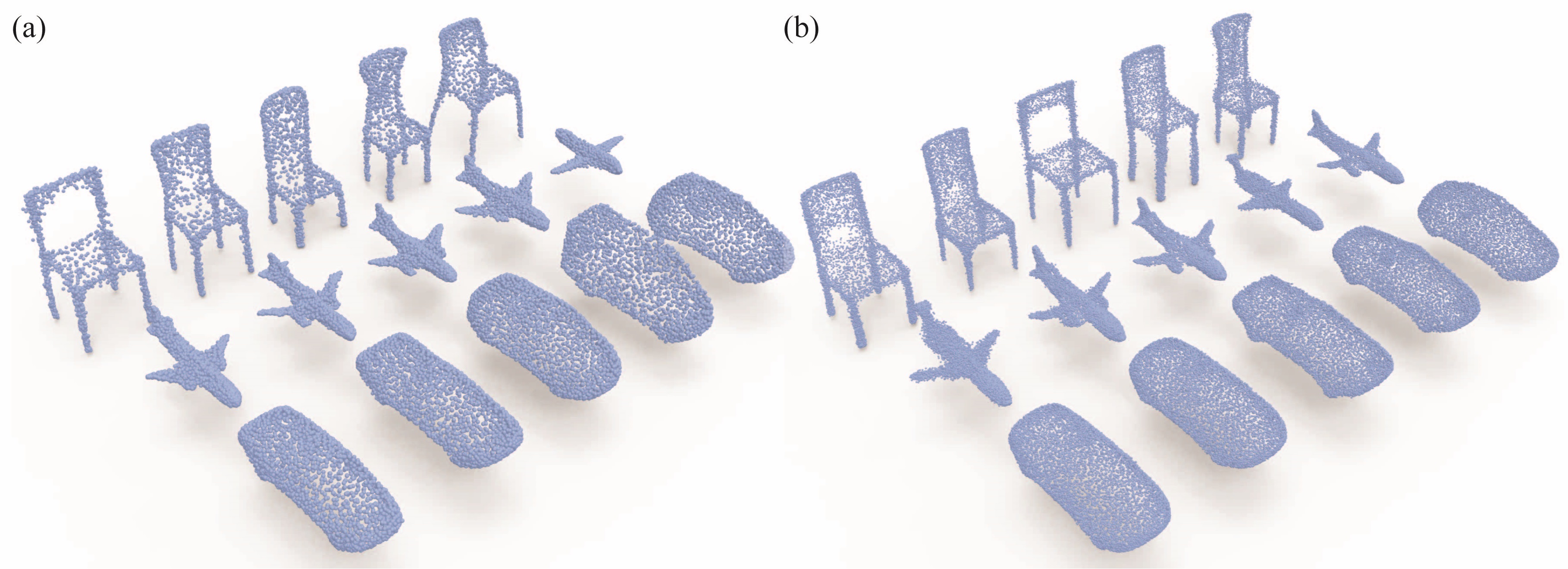}
  %
  %
\caption{\label{fig:MultiRes} Randomly selected shapes with $2000$ points in (a) and $8000$ points in (b) from reconstructed shapes producing Table~\ref{table:t10}.}
\end{figure}

Fig.~\ref{fig:PCDRComp} shows a qualitative comparison with the point cloud renderers used in CAP and DPC at a resolution of 2000 points. We find that DRWR generalizes better for rarely seen shapes and achieves higher accuracies by more uniformly distributing the recovered 3D points. In addition, we visualize more high fidelity shapes at resolutions of 2000, 8000, and 16000 points in Fig.~\ref{fig:MultiRes} (a), (b) and Fig.~\ref{fig:Vis_128}, respectively. We find that DRWR can train networks to generate plausible point clouds with different numbers of points from images, while DRWR would recover more geometry details when using more points to represent a shape.


\begin{figure}[tb]
  \centering
   \includegraphics[width=\linewidth]{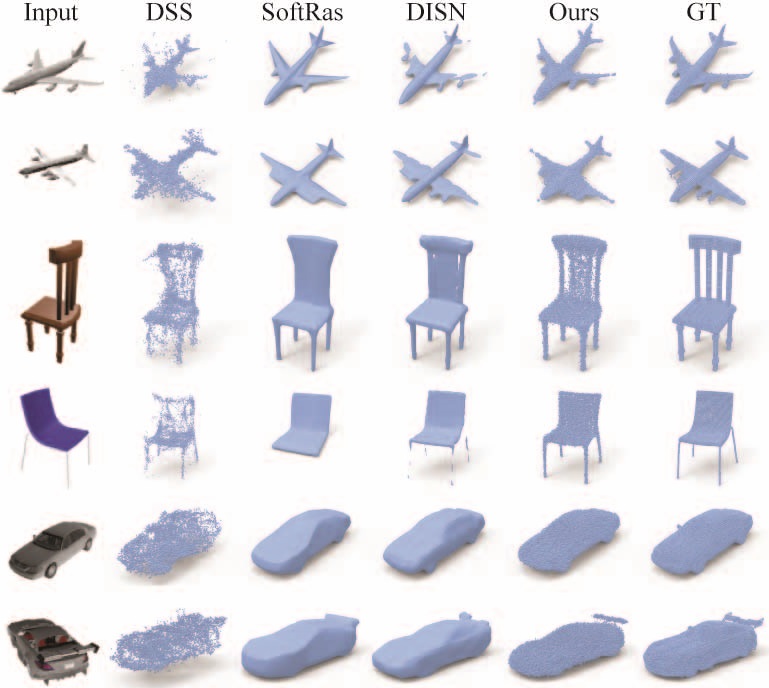}
  %
  %
\caption{\label{fig:DRComp} Qualitative comparison with differentiable renderers for different 3D representations and supervised learning methods.}
\end{figure}


\begin{figure*}[tb]
  \centering
   \includegraphics[width=\linewidth]{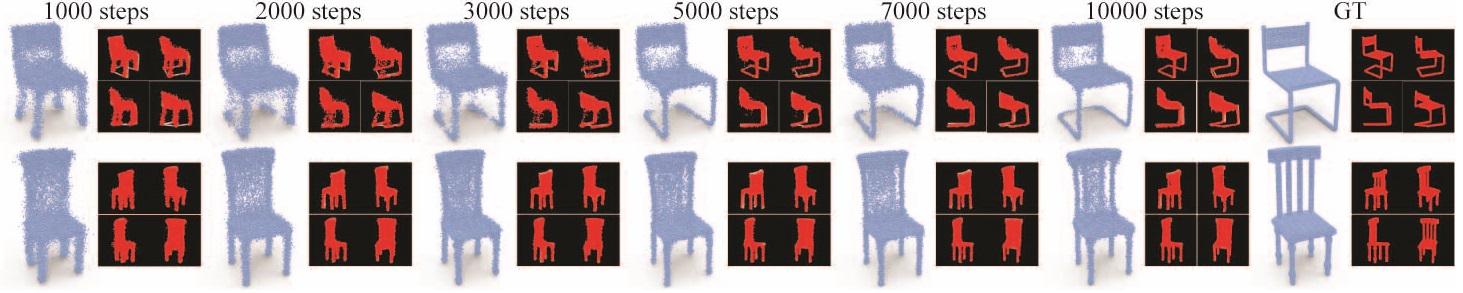}
  %
  %
\caption{\label{fig:ChangingPattern} A shape reconstructed from a test image using network parameters in different steps.}
\end{figure*}


\begin{table*}[ht]
\centering
\caption{Quantitative comparison with differentiable renderers for different 3D representations and supervised methods in terms of IoU.}  
\resizebox{\linewidth}{!}{
    \begin{tabular}{|c|c|c|c|c|c||c|c|c|c|c|c|c|c|c|}  
     \hline
     &\multicolumn{5}{|c||}{Unsupervised differentiable renderers} &\multicolumn{7}{|c|}{Supervised structure learning methods}& \\
     \hline
          & PTN & NMR & SoftR&DIB-R& Ours&DISN&OccNet&IMNET&3DN&Pix2Mesh&R2N2&AtlasNet&Ours\\ 
     \hline

     Car &71.2&71.3&77.1&\textbf{78.8}&75.3&74.3&73.7&74.5&59.4&50.1&66.1&22.0&\textbf{75.3}\\
     Plane&55.6&58.5&58.4&57.0&\textbf{62.2}&57.5&57.1&55.4&54.3&51.5&42.6&39.2&\textbf{62.2}\\
     Chair&44.9&41.4&49.7&52.7&\textbf{58.1}&54.3&50.1&52.2&34.4&40.2&43.9&25.7&\textbf{58.1}\\
     \hline
     Mean&57.2&57.1&61.7&62.8&\textbf{65.2}&62.0&60.3&60.7&49.4&47.3&50.9&29.0&\textbf{65.2}\\
     \hline
   \end{tabular}}
   \label{table:VOXEL}
\end{table*}

\noindent\textbf{IoU comparisons. }We further compare DRWR in terms of IoU with rendering-based differentiable renderers for other 3D representations, such as meshes and voxel grids. The comparison includes Perspective Transform Nets (PTN)~\cite{YanNIPS2016}, Neural Mesh Renderer (NMR)~\cite{KatoUH18}, SoftRasterizer (SoftR)~\cite{liu2019softras}, and Interpolation-based Differentiable Renderer (DIB-R)~\cite{DBLP:journals/corr/abs-1908-01210}. The former two methods are voxel-based, while the latter two methods are mesh-based. We report the results of NMR, SoftR and DIB-R from ~\cite{DBLP:journals/corr/abs-1908-01210}. To produce our IoU, we voxelize the point clouds predicted from images with a resolution of $128^2$ in Table~\ref{table:t10} into voxel grids with resolution $32^3$ to compare to the same ground truth as other methods.

The quantitative comparison is shown in the ``unsupervised'' part of Table~\ref{table:VOXEL}. Our results significantly outperform the state-of-the-art differentiable renderers in terms of mean IoU, where we achieve the best under airplanes and chairs. For cars, our results are better than voxel-based renderers but only comparable to mesh-based renderers. This is because the mesh-based renderers are good at representing large areas of flat surfaces, such as cars. Mesh-based approaches are limited to a fixed (usually spherical) mesh topology, however. This leads to inaccuracies when representing more complex surfaces, such as chairs, which often exhibit non-spherical topology.

We further conduct qualitative comparisons with the latest differentiable renderers including point cloud-based DSS~\cite{Yifan:DSS:2019} and mesh-based SoftR in Fig.~\ref{fig:DRComp}. For fair comparison, we also produce results of DSS from four views which keeps the same as DRWR in training, where DSS uses its own camera system to generate rendered images from ground truth point clouds. In addition, we show our results at a resolution of 16000 points and ground truth involved in Table~\ref{table:t10}. Compared to DSS and SoftR, DRWR can reveal more geometry details.


\noindent\textbf{Supervised methods. }Finally, we compare DRWR with the latest supervised 3D structure learning methods. In the first experiment, we compare with NOX~\cite{NIPS2019_Srinath} for point clouds, where we report our results using the evaluation code released by NOX~\cite{NIPS2019_Srinath}. Following the same setting, we employ the point clouds reconstructed from input images with a resolution of $64^2$ in Table~\ref{table:t10}, scale each predicted point cloud such that the diagonal of its bounding box is one, and resample a ground truth point cloud to $8000$ points if there are more than $8000$ points in it. Table~\ref{table:NOX} shows that our results significantly outperform NOX under all three classes.

The ``supervised'' part in Table~\ref{table:VOXEL} reports the numerical comparison with the state-of-the-art supervised methods, where we used the same approach as before to obtain our IoU results. We significantly outperform supervised methods under all three shape categories. Fig.~\ref{fig:DRComp} also shows a visual comparison with DISN, the best supervised method, illustrating that we obtain similar quality.



\begin{table}[h]
\centering
\caption{Comparison (CD) with latest supervised method NOX.}  
\resizebox{0.6\linewidth}{!}{
    \begin{tabular}{|c|c|c|c|}  
     \hline
          & Cars & Airplanes & Chairs\\   
     \hline
       NOX& 0.3331 & 0.2795 & 0.4637 \\
       Ours & \textbf{0.0446} & \textbf{0.0527} & \textbf{0.0540} \\
     \hline
   \end{tabular}}
   \label{table:NOX}
\end{table}


\begin{figure*}[!t]
  \centering
   \includegraphics[width=\linewidth]{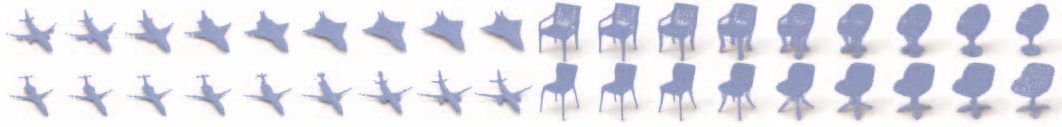}
  %
  %
\caption{\label{fig:Interpolation} Interpolated shapes in the learned 3D feature space.}
\end{figure*}

\subsection{Ablation studies and analysis}

\noindent\textbf{Ablation studies. }We conduct ablation studies to justify the effectiveness of each element in DRWR under airplanes at a resolution of 2000 points in Table.~\ref{table:t10}. In Table~\ref{table:ablation}, we report our results with only the smooth silhouette loss in Eq.~(\ref{eq:unary1}) (``$l_1$''), with only structure-aware repulsion loss in Eq.~(\ref{eq:pairwise}) (``$l_2$''), with binary pixel loss in Eq.~(\ref{eq:unary}) and structure-aware repulsion loss (``Pixel+$l_2$''), with smooth silhouette pixel loss in Eq.~(\ref{eq:unary1}) and repulsion loss without indicator weights (``$l_1$+no $w_j^i$''), and with smooth silhouette loss in Eq.~(\ref{eq:unary1}) and repulsion loss without boundary bias (``$l_1$+no $\delta_j^i$''). We also report results with fewer views of each shape in training, such as $I=2$ and $I=3$ views.

\begin{table}[h]
\centering
\caption{Ablation studies in terms of CD.}  
\resizebox{\linewidth}{!}{
    \begin{tabular}{|c|c|c|c|c|c||c|c||c|}  
     \hline
          & $l_1$ & $l_2$ & Pixel+$l_2$ &$l_1$+no $w_j^i$&$l_1$+no $\delta_j^i$&$I=2$&$I=3$&$I=4$\\   
     \hline
       CD &19.50&139.10&24.59&4.58&4.41&4.79&4.54&\textbf{4.01}\\
     \hline
   \end{tabular}}
   \label{table:ablation}
\end{table}

The ablation studies show that DRWR cannot learn the structure of shapes using only $l_1$ or $l_2$, nor without the smooth silhouette loss because of the local minimum issue. The structure awareness brought by indicator weights and boundary bias also contributes to the reconstruction accuracy and optimization efficiency, which is also demonstrated by the loss $L$ comparison in the first 10000 steps in Fig.~\ref{fig:final1}. The loss comparison shows that the structure awareness significantly decreases the conflict with the smooth silhouette loss, which leads to lower loss and faster convergence. In addition, using fewer views than our $I=4$ views in training also degenerates the structure learning performance.

\begin{table}[h]
\centering
\caption{Efficiency comparison in terms of training time.}  
\resizebox{\linewidth}{!}{
 \begin{tabular}{|c|c|c|c|c|c|}  
     \hline
          &\multirow{3}{*}{Modality}&\multirow{3}{*}{Rendering}& $32^2$ image& $64^2$ image & $128^2$ image\\   
          &&&2000 points/&8000 points/&16000 points/\\
          &&&$32^3$ voxels&$64^3$ voxels&$128^3$ voxels\\
     \hline
       DRC&Voxel&Yes& $\approx$14h& $\approx$60h& Out of memory \\
       DPC&Points&Yes& $\approx$14h & $\approx$24h & $\approx$72h \\
       \hline
       Ours &Points&No& $\approx$\textbf{7}h & $\approx$\textbf{12}h & $\approx$\textbf{36}h \\
     \hline
   \end{tabular}}
   \label{table:direction}
\end{table}

\noindent\textbf{Efficiency. }We highlight the efficiency of DRWR by comparing our network training time with state-of-the-art differentiable renderers for 3D shapes. The voxel-based method of DRC~\cite{TulsianiZEM17} suffers from a huge computation burden due to the cubic complexity of voxel grids, which limits it to work only in low resolutions such as $32^3$ and $64^3$ with slow convergence. Although the point cloud-based method of DPC~\cite{InsafutdinovD18} does not require 3D convolution layers as DRC, which improves the efficiency and enables to work in higher resolution, the rendering procedure still requires intensive computation with discrete 3D grids. Therefore, DPC requires more time ($6\times10^{5}$ mini-batch iterations) during training than DRWR ($1\times10^{5}$ mini-batch iterations), thanks to the removal of rendering.



\noindent\textbf{Optimization. }We visualize the optimization process in Fig.~\ref{fig:ChangingPattern}. We use the parameters learned in different steps during training to reconstruct a shape from a corresponding image in the test set. In addition, we show the 2D projections on four views. Using an image from the test rather than training set can better demonstrate the generalization ability learned in optimization, justifying our effectiveness.

\begin{figure}[tb]
  \centering
   \includegraphics[width=\linewidth]{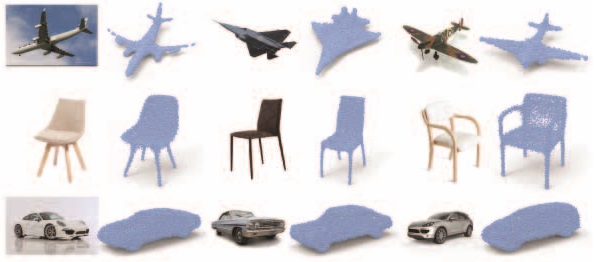}
  %
  %
\caption{\label{fig:RealImg} Qualitative demonstration of shapes reconstructed from real images.}
\end{figure}

\noindent\textbf{Adaptation to real images. }We evaluate the adaptation ability to real images in the network trained using DRWR in Fig.~\ref{fig:RealImg}. Using the parameters learned in Table~\ref{table:t10}, we reconstruct shapes at a resolution of 16000 points from real images selected from the Internet. The high fidelity of reconstructed shapes demonstrates that DRWR can train networks to adapt to real images very well.


\noindent\textbf{Latent space. }We visualize the latent space learned in the network that we train using DRWR. We employ a trained network to reconstruct shapes using 1024-dimensional latent codes that are interpolated from two known codes of two shapes. As shown in Fig.~\ref{fig:Interpolation}, the interpolated shapes in the smooth transformation show that DRWR helps the network to learn a meaningful latent space.

\section{Conclusion}
We propose DRWR for unsupervised 3D structure learning using point clouds. DRWR successfully removes the rendering step that is commonly used in state-of-the-art differentiable renderers. While rendering requires additional computation, a key observation from our experiments is that it does not contribute to improving accuracy in 3D structure learning. DRWR achieves this by minimizing a unary and a pairwise loss. The unary loss uses a smooth silhouette loss to pull all projections into the foreground by effectively resolving the severe local minimum issue, while the pairwise loss uses structure-aware repulsion to efficiently push pairs of projections in the foreground away from each other by adaptively weighting the repulsion according to the 2D structure. The effectiveness of DRWR is justified by superior experimental results over the state-of-the-art.

\section*{Acknowledgements}
We thank the anonymous reviewers for reviewing our paper and providing helpful comments. This work was supported by National Key R\&D Program of China (2018YFB0505400), in part by Tsinghua-Kuaishou Institute of Future Media Data, and NSF (award 1813583).

\nocite{langley00}

\bibliography{../../paper}
\bibliographystyle{icml2020}

%
%
%

\end{document}